\documentclass{article}

\PassOptionsToPackage{numbers, compress}{natbib}

\usepackage[preprint]{neurips_2024}

\usepackage[utf8]{inputenc} 
\usepackage[T1]{fontenc}    
\usepackage{hyperref}       
\usepackage{url}            
\usepackage{booktabs}       
\usepackage{amsfonts}       
\usepackage{nicefrac}       
\usepackage{microtype}      
\usepackage{xcolor}         

\usepackage{graphicx}
\usepackage{booktabs}
\usepackage{multirow}
\usepackage{amsmath}
\usepackage{xspace}
\usepackage{colortbl}
\usepackage{wrapfig}

\title{PLUG: Revisiting Amodal Segmentation with Foundation Model and Hierarchical Focus}

\author{%
Zhaochen Liu\textsuperscript{\rm 1,2}, Limeng Qiao\textsuperscript{\rm 3}, Xiangxiang Chu\textsuperscript{\rm 3}, Tingting Jiang\textsuperscript{\rm 1,4}\thanks{Corresponding author.}\\
\textsuperscript{\rm 1}National Engineering Research Center of Visual Technology,\\ National Key Laboratory for Multimedia Information Processing, \\ School of Computer Science, Peking University \\
\textsuperscript{\rm 2}AI Innovation Center, School of Computer Science, Peking University \\  
\textsuperscript{\rm 3}Meituan Inc. \\ 
\textsuperscript{\rm 4}National Biomedical Imaging Center, Peking University\\ 
\texttt{\{dreamerliu, qiaolm, ttjiang\}@pku.edu.cn, cxxgtxy@gmail.com}
}

\begin{document}

\maketitle

\begin{abstract}
  Aiming to predict the complete shapes of partially occluded objects, amodal segmentation is an important step towards visual intelligence. With crucial significance, practical prior knowledge derives from sufficient training, while limited amodal annotations pose challenges to achieve better performance. To tackle this problem, utilizing the mighty priors accumulated in the foundation model, we propose the first SAM-based amodal segmentation approach, PLUG. Methodologically, a novel framework with hierarchical focus is presented to better adapt the task characteristics and unleash the potential capabilities of SAM. In the region level, due to the association and division in visible and occluded areas, inmodal and amodal regions are assigned as the focuses of distinct branches to avoid mutual disturbance. In the point level, we introduce the concept of uncertainty to explicitly assist the model in identifying and focusing on ambiguous points. Guided by the uncertainty map, a computation-economic point loss is applied to improve the accuracy of predicted boundaries. Experiments are conducted on several prominent datasets, and the results show that our proposed method outperforms existing methods with large margins. Even with fewer total parameters, our method still exhibits remarkable advantages. The code and models will be publicly available.
\end{abstract}

\section{Introduction}
\label{sec:intro}
Occlusion is a common phenomenon in the real world, thus comprehending occlusion is a vital capability of visual intelligence~\cite{palmer1999vision}. In computer vision, amodal segmentation was proposed in 2016~\cite{li2016amodal} to predict the complete shape of an object containing both the visible portion and the occluded portion. Amodal segmentation has been widely studied since then. Researchers explored fully supervised approaches~\cite{li2016amodal, zhu2017semantic, qi2019amodal, wang2020robust, xiao2021amodal, li20222d, gao2023coarse}, weakly supervised approaches~\cite{zhan2020self, nguyen2021weakly, sun2022amodal, liu2024blade}, and diverse applications including self-driving~\cite{qi2019amodal, breitenstein2022amodal}, image augmentation~\cite{gkitsas2021panodr, li20222d, pintore2022instant} and robotic gripping systems~\cite{wada2018instance, wada2019joint, inagaki2019detecting}.

Amodal segmentation performance is closely related to the implicit prior knowledge of the model~\cite{ao2023image}. Despite the achieved progress, the pursuit of high quality priors raises a demand for sufficient training, while large-scale amodal annotations are time-consuming and labor-intensive~\cite{liu2024blade}. Encouragingly, foundation models emerge and develop rapidly in recent years~\cite{zhou2023comprehensive, awais2023foundational}. Pre-training through massive data brings a broad and effective cognition to these foundation models, thereby inspiring a foundation-model-based paradigm of algorithm design. The Segment Anything Model, SAM~\cite{kirillov2023segment}, is a typical and influential foundation model for segmentation. Is it feasible to utilize the potential of SAM to further improve the amodal segmentation performance without introducing extra data? 

To reach this goal, we propose PLUG, the first SAM-based amodal segmentation approach. 
Naturally, limited accessable data leads us towards parameter-efficient finetuning, while achieving competent training is still challenging due to the complexity of amodal segmentation. 
Observing the positive impact of emphasizing highlights in human learning process, we expect to facilitate the model training by establishing analogous focused learning. Therefore, an elaborately-designed framework with hierarchical focus appropriate to the task characteristics is applied to better deliver the advantages of SAM, in which focused regions and points are designated explicitly.


In the region level, we realize that though related to each other, visible area segmentation and occluded area segmentation actually belong to different tasks. The former mainly focuses on the information \emph{within} the region, while the latter requires reasonable inference considering the \emph{entire} object region. 
However, the principle of parameter-efficient finetuning is to ingeniously decrease desired trainable parameters through targeting \emph{one} specific task~\cite{aghajanyan2020intrinsic,hu2022lora}. Therefore, there is a contradiction when applying existing methods directly on amodal segmentation. To address this issue, we introduce a foundation-model-applicable parallel structure to separately focus on the inmodal and amodal regions, which involves two independent branches with different adapters to ensure mutual disturbance is avoided. Subsequently, the predictions from these two branches are fused and rectified in a simple yet effective refine module, thus retaining the relevance and close collaboration.

In the point level, we notice the significance of different points is \emph{not} uniform for the shape of some object, among which points near the edge reveal more impact. Since the boundary prediction counts, we conceive a design to particularly focus on pertinent points thereby supporting promoted accuracy. Specifically, we introduce the concept of uncertainty based on the cross-entropy that originally possesses relevant meanings. Guided by the uncertainty map, a supplemental point loss is employed in the training. Inspired by PointRend~\cite{kirillov2020pointrend}, we combine random sampling and sorted selection to find points with high uncertainty without excessive additional computation. Intuitively, points near the boundary will exactly be picked due to ambiguous prediction. These points obtain enhanced penalty by the point loss. Moreover, to further prompt the focus on these points, the uncertainty is also explicitly involved in the refine module to provide practical clues for the fusion process.


\begin{figure}[t]
  \centering
   \includegraphics[width=0.72\linewidth]{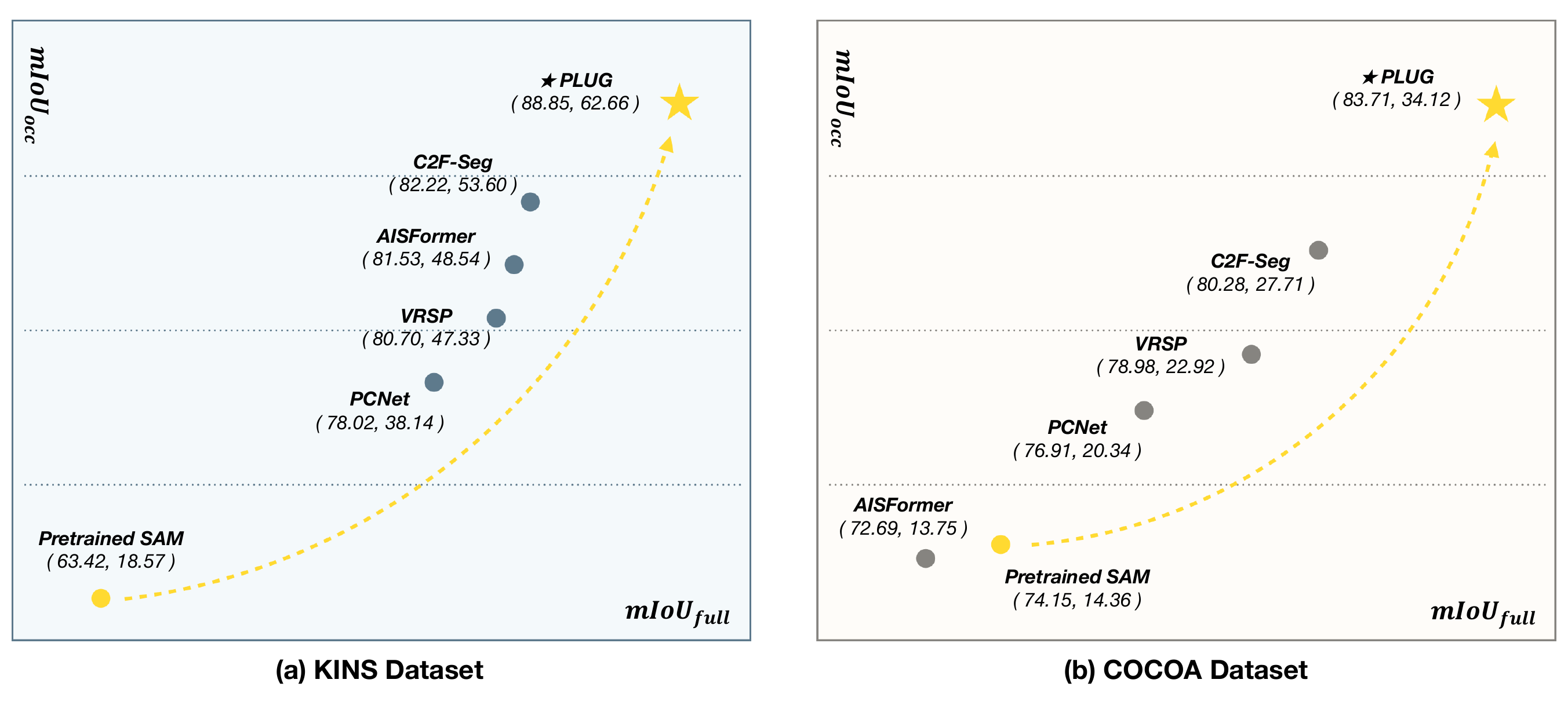}
   \caption{\textbf{The visualization of performance comparison.} 
   $\text{mIoU}_{full}, \text{mIoU}_{occ}$ represent the mean IoU for the complete mask and the occluded region of each object, respectively.}
   \label{fig:performance}
\end{figure}

To evaluate the performance of our proposed method, we conduct experiments on the prominent KINS dataset~\cite{qi2019amodal} and COCOA dataset~\cite{follmann2019learning}. Our proposed method is implemented grounded on the advanced LoRA~\cite{hu2022lora}, low-rank adaptation structure that exhibits wide applicability in parameter-efficient finetuning. As shown in Fig.~\ref{fig:performance}, PLUG attains a remarkable upgradation in amodal segmentation compared to the pretrained SAM, and outperforms all existing approaches with large margins achieving state-of-the-art performance on both datasets. Even with fewer total parameters, PLUG still displays notable advantages. Our contributions can be summarized as follows:

\begin{itemize}
\item We exploit the foundation model to surmount the data-deficient issue in amodal segmentation task, and propose the first SAM-based amodal segmentation approach.
\item According to the task characteristics, a novel framework with hierarchical focus is finely designed containing \textbf{P}arallel \textbf{L}oRA and \textbf{U}ncertainty \textbf{G}uidance~(\textbf{PLUG}), with which the model can efficiently focus on specific regions and ambiguous points thus improving the training effectiveness and unleashing the distinguished capabilities of the foundation model.
\item Our proposed method significantly outperforms existing approaches with large margins, reaching state-of-the-art performance.
\end{itemize}

\section{Related Work}
\label{sec:related}
\paragraph{Amodal Segmentation.}
The goal of amodal segmentation is to predict the complete shape of the occluded object through the deep model, which is essential to achieve visual intelligence. In the early stages of research, besides direct approach~\cite{li2016amodal, zhu2017semantic, qi2019amodal}, many fully supervised methods have been proposed with various concepts involved, such as depth relationship~\cite{zhang2019learning}, region correlation~\cite{follmann2019learning,ke2021deep}, and shape priors~\cite{xiao2021amodal,li20222d,chen2023amodal,gao2023coarse}. Immediately afterward, a series of weakly supervised methods~\cite{zhan2020self,nguyen2021weakly,kortylewski2020compositional,kortylewski2021compositional,sun2022amodal,liu2024blade} begin to appear with the supervision of simpler annotations, such as bounding boxes and categories. Exploiting the capabilities of amodal segmentation, researchers are concerned with its diverse applications. Amodal segmentation is critical to guarantee operational safety and reliability of various intelligent systems, such as autonomous driving~\cite{qi2019amodal, breitenstein2022amodal} and robotic grasping~\cite{wada2018instance, wada2019joint, inagaki2019detecting}, while also forwarding image augmentation~\cite{gkitsas2021panodr, li20222d, pintore2022instant}. Though numerous related work is delivered, some issues still exist in this field. The aforementioned models exhibit a tendency towards complexity and face the challenge of universality, which is not friendly to practical use. Benefiting from the versatile feature extraction capabilities provided by large-scale vision foundational models, our method introduces a brand-new foundation-model-based architecture to tackle these problems.

\paragraph{Foundation Models.}
Recently, remarkable progress has been made in developing visual foundation models~(VFMs). Trained on large-scale data to acquire foundational capabilities, these models can be adapted to a wide range of related downstream vision tasks. Vision-language models like CLIP~\cite{radford2021learning} are first studied, which demonstrate promising zero-shot generalization performance on different tasks. Following this paradigm, a series of models are proposed to gradually improve generalization and versatility~\cite{fang2023eva,yao2021filip,li2022blip,li2022grounded,cherti2023reproducible,liu2023grounding} and to expand applicable fields~\cite{luddecke2022image,wang2023seggpt,kirillov2023segment,zou2024segment,zhang2023faster,wang2023caption}. Among them, the Segment Anything Model, SAM~\cite{kirillov2023segment}, is a classic work, which performs a class-agnostic segmentation given an image and a visual prompt such as box, point, or mask. By training on a model-in-the-loop dataset with billions of object masks, SAM shows excellent performance on various segmentation tasks~\cite{awais2023foundational}. While the recent surge of the visual foundation models, in the field of amodal segmentation research, the practical exploration of using these universal VLMs is still few.

\paragraph{Parameter-efficient Fine-tuning.}
Due to the massively scaling up of both data and model size, the training of almost all current VFMs is computation-intensive and storage-consuming. A compelling solution for this challenge is parameter-efficient fine-tuning, which is to finetune only a small set of external parameters rather than the entire model while still capitalizing on the outstanding capabilities of foundation models~\cite{hu2023llm}. To achieve this, diversified  approaches have been proposed. In the prompt tuning~\cite{lester2021power}, a trainable tensor is added as a prefix to the input embeddings. The Series Adapter~\cite{wang2022adamix,houlsby2019parameter,karimi2021compacter} proposes the insight of incorporating additional learnable modules in a sequential manner within a specific sub-layer. The reparametrization-based methods\cite{aghajanyan2020intrinsic,hu2022lora,edalati2022krona} transform network weights using low-rank techniques.  Currently, LoRA\cite{hu2022lora} is widely used, which introduces rank decomposition matrices for specific tasks. Although LoRA effectively reduces the number of trainable parameters while retaining the performance, the original design is still not satisfactory enough for amodal segmentation. To address this problem, we propose a novel PLUG framework in this paper.

\section{Method}
\label{sec:method}

\subsection{Task Definition}
For a partially occluded object, amodal segmentation is to segment its complete shape containing both visible and occluded areas. Unlike some other methods~\cite{zhan2020self, gao2023coarse}, our PLUG do \emph{not} require the visible mask as input. For each object $i$, we take only the original image $\mathbf{I}$ and the bounding box $\mathbf{B}_v^i$ of the visible portion that indicates the region of interest as the input. The final output is the predicted binary amodal mask $\Tilde{\mathbf{M}}_a^i$. As for the supervision signal, we adopt the ground-truth binary visible mask $\mathbf{M}_v^i$ and amodal mask $\mathbf{M}_a^i$. Category labels are not utilized in the training.

\subsection{Overall Architecture}
\begin{figure}[t]
  \centering
   \includegraphics[width=\linewidth]{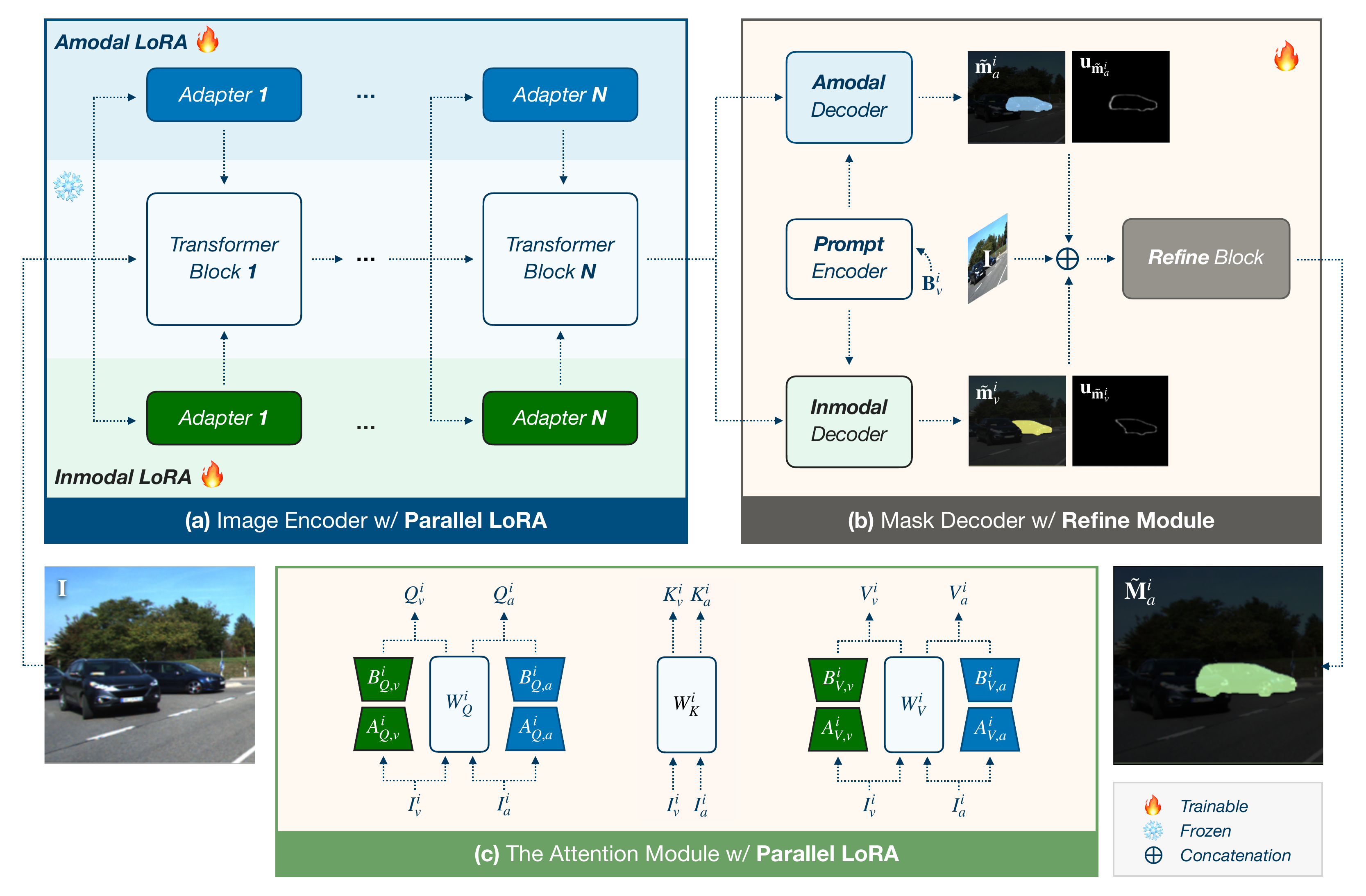}
   \caption{\textbf{The architecture of PLUG.} (a,b) On the basis of SAM, two parallel sets of LoRA adapters~(Inmodal LoRA, Amodal LoRA) and corresponding two mask decoders~(Inmodal Decoder, Amodal Decoder) are introduced to separately process diverse regions and avoid mutual disturbance. Guided by uncertainty maps~(defined in Sec.~\ref{sec:point}), a simple yet effective refine module is added afterwards to rectify ambiguous points near the boundary. The refine module takes the original image, the coarse predictions and the uncertainty maps as input. (c) In each transformer block of the image encoder, low-rank adaptation matrices are applied to the attention module. The calculation of $Q,V$ passes two parallel side roads focusing on inmodal and amodal regions respectively~(refer to Sec.~\ref{sec:parallel}).}
   \label{fig:plug-framework}
\end{figure}

As shown in Fig.~\ref{fig:plug-framework}, our proposed approach is developed on the foundation segmentation model SAM~\cite{kirillov2023segment}. The architecture can be divided into four parts: image encoder, prompt encoder, mask decoder, and refine module.
\textbf{(1)~Image Encoder.} We freeze the image encoder in SAM that contains most parameters of the pretrained model~(632M/636M for the ViT-H~\cite{dosovitskiy2021an} version), and utilize low-rank adaptation~\cite{hu2022lora} to finetune the encoder. Introducing trainable rank decomposition matrices into each layer of the transformer architecture, LoRA can substantially limit the number of trainable parameters. Here we inject two parallel branches of LoRA, namely Inmodal LoRA and Amodal LoRA. Through the modified image encoder, inmodal image embeddings $\mathbf{E}_{v}$ and amodal image embeddings $\mathbf{E}_{a}$ are extracted from the original image $\mathbf{I}$.
\textbf{(2)~Prompt Encoder.} The trainable prompt encoder is used to convert the input bounding box $\mathbf{B}_v^i$ into corresponding prompt embeddings $\mathbf{E}_{p}^i$.
\textbf{(3)~Mask Decoder.} We employ two trainable mask decoders, namely Inmodal Decoder and Amodal Decoder, for the two branches. The embeddings $\mathbf{E}_{v},\mathbf{E}_{p}^i$ and $\mathbf{E}_{a},\mathbf{E}_{p}^i$ are exploited respectively to produce coarse inmodal prediction $\Tilde{\mathbf{m}}_v^i$ and coarse amodal prediction $\Tilde{\mathbf{m}}_a^i$ by these mask decoders.
\textbf{(4)~Refine Module.} Coarse predictions $\Tilde{\mathbf{m}}_v^i, \Tilde{\mathbf{m}}_a^i$ from the two separate branches are integrated and refined in the subsequent refine module. The refined amodal mask $\Tilde{\mathbf{M}}_a^i$ is adopted as the final output.
In summary, the trainable components in the network are the parallel LoRAs, the prompt encoder, the mask decoders, and the refine module. Therefore, the overall number of trainable parameters is effectively limited.

\subsection{Region-level Focus: Parallel LoRA}
\label{sec:parallel}
In experiments, we find that the prediction quality is not satisfactory enough when we directly use LoRA to finetune the SAM model and predict the amodal mask $\Tilde{\mathbf{M}}_a^i$. We realize this plain approach may contradict LoRA's original intention. Obviously, for some object $i$, the complete mask of $\mathbf{M}_a^i$ can be divided into the visible segment $\mathbf{M}_v^i$ and the occluded segment $\mathbf{M}_o^i$:
\begin{equation}
\label{eq:dcp1}
    \mathbf{M}_a^i = \mathbf{M}_v^i + \mathbf{M}_o^i.
\end{equation}
The insight of LoRA is that for a specific task, the updates to the weights of pretrained model have a low intrinsic rank~\cite{hu2022lora}, while the segmentation of visible areas and occluded areas can be regarded as two distinct tasks. Though related, visible portion segmentation and occluded portion segmentation are diverse in characteristics. Firstly, the segmentation of $\mathbf{M}_v^i$ mainly focuses on the information within the region $\mathbf{B}_v^i$, while the segmentation of the $\mathbf{M}_o^i$ requires information from the entire object region. Secondly, the segmentation of the $\mathbf{M}_o^i$ involves moderate inference utilizing the implicit shape priors in the model. Therefore, $\mathbf{M}_v^i$ and $\mathbf{M}_o^i$ should be processed separately to avoid mutual disturbance. Observing this, we introduce the parallel LoRA structure. Lacking sufficient semantics while completely differing from regular \emph{modal} segmentation, it is not suitable to directly build a branch for the occluded part $\mathbf{M}_o^i$. Thus, we rewrite Eq.~\ref{eq:dcp1} as
\begin{equation}
\label{eq:dcp2}
    \mathbf{M}_a^i = \mathbf{M}_v^i + (\mathbf{M}_a^i - \mathbf{M}_v^i),
\end{equation}
where only $\mathbf{M}_v^i$ and $\mathbf{M}_a^i$ are involved. Guided by Eq.~\ref{eq:dcp2}, we alternatively adopt a coarse visible branch and a coarse amodal branch to predict the coarse visible mask $\Tilde{\mathbf{m}}_v^i$ and the coarse amodal mask $\Tilde{\mathbf{m}}_a^i$, respectively. $\Tilde{\mathbf{m}}_v^i, \Tilde{\mathbf{m}}_a^i$ are fused by a subsequent refine module to obtain the final output $\Tilde{\mathbf{M}}_a^i$.


Specifically, we adopt two parallel sets of adapters~(Inmodal LoRA $L_v$, Amodal LoRA $L_a$) for the image encoder and corresponding two mask decoders~(Inmodal Decoder $D_v$, Amodal Decoder $D_a$) as shown in Fig.~\ref{fig:plug-framework}. The LoRA adapters are added to each transformer block in the image encoder and initialized randomly. With parallel LoRA, each transformer block has two side roads in the attention module as shown in Fig.~\ref{fig:plug-framework}. More detailed, in the $i$-th transformer block, LoRA is applied to the attention weight matrices $W_Q^i, W_V^i$. For $W_Q^i$, we have inmodal adaptation matrices $A_{Q,v}^i,B_{Q,v}^i$ and amodal adaptation matrices $A_{Q,a}^i,B_{Q,a}^i$:
\begin{equation}
    Q_{\{v,a\}}^i=W_Q^iI_{\{v,a\}}^i+B_{Q,\{v,a\}}^iA_{Q,\{v,a\}}^iI_{\{v,a\}}^i.
\end{equation}
Similarly, for $W_V^i$ we have $A_{V,v}^i,B_{V,v}^i,A_{V,a}^i,B_{V,a}^i$:
\begin{equation}
    V_{\{v,a\}}^i=W_V^iI_{\{v,a\}}^i+B_{V,\{v,a\}}^iA_{V,\{v,a\}}^iI_{\{v,a\}}^i.
\end{equation}
In the above expressions, $v,a$ represent ``visible'' and ``amodal'', respectively. $I_v^i, I_a^i$ are the input of this attention module from the inmodal branch and the amodal branch before.

As for the mask decoders, Inmodal Decoder processes inmodal image embeddings and prompt embeddings, while Amodal Decoder processes amodal image embeddings and prompt embeddings:
\begin{equation}
    \Tilde{\mathbf{m}}_v^i=D_v(\mathbf{E}_{v},\mathbf{E}_{p}^i),
    \Tilde{\mathbf{m}}_a^i=D_a(\mathbf{E}_{a},\mathbf{E}_{p}^i).
\end{equation}
Both decoders are initialized with the same parameters as the mask decoder in the pretrained SAM. All trainable components in the architecture are trained concurrently to ensure close collaboration.

\subsection{Point-level Focus: Uncertainty Guidance}
\label{sec:point}
\begin{figure}[t]
  \centering
   \includegraphics[width=\linewidth]{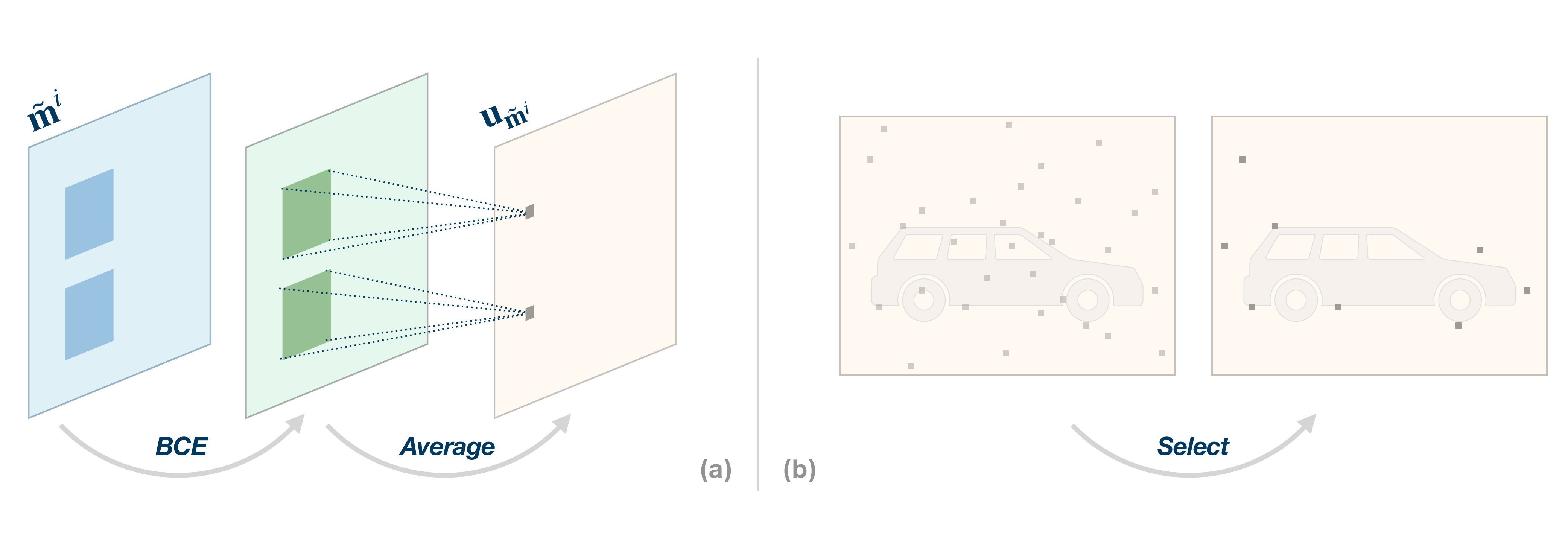}
   \caption{\textbf{An illustration of the uncertainty guidance.} (a) The uncertainty of each pixel is defined as the average cross entropy of its neighborhood. (b) We select $cK$ points with high uncertainty~(top $cK$) and stochastic $(1-c)K$ points from randomly chosen $nK$ points to apply the point loss.}
   \label{fig:point}
\end{figure}

In the segmentation task, we actually focus on the points near the boundary of the object, which determine the shape to some extent. That is, the significance of different points is diverse, among which points near the edge usually possess more importance. Accordingly, these points are also more challenging in the prediction. We hope to reflect the imbalanced importance in our method and help identify these key points more quickly. For this reason, we introduce the concept of uncertainty. As shown in Fig.~\ref{fig:point}, we define the uncertainty of some pixel as the \emph{average} cross-entropy within the neighborhood. For point $\mathbf{x}=(x_1,x_2)$ in predicted logits $\mathbf{m}$, the uncertainty is
\begin{equation}
    u_{\mathbf{m}}(\mathbf{x}) = \scalebox{1.2}{\(\sum\nolimits\)}_{(i,j)\in N(\mathbf{x},\epsilon)}\ (2\epsilon-1)^{-2}\ \mathrm{CE}(\mathbf{m}(i,j)),
\end{equation}
where $\epsilon$ is a constant parameter indicates the neighborhood size. 
Any point $(i,j)$ in $N(\mathbf{x},\epsilon)$ satisfies
\begin{equation}
    |i-x_1|<\epsilon \wedge |j-x_2|<\epsilon.
\end{equation}
Cross-entropy succinctly measures the chaos degree of predicted logits. If a point is located in a neighborhood with relatively chaotic predictions, we can assert that its uncertainty is high. For our issue, the cross entropy in the formula is actually binary cross entropy:
\begin{equation}
    \mathrm{CE}(\mathbf{m}(i,j))=-[\mathbf{m}(i,j)\log\mathbf{m}(i,j)+(1-\mathbf{m}(i,j))\log(1-\mathbf{m}(i,j))].
\end{equation}
By sequentially calculating the uncertainty of each point, we obtain an uncertainty map $\mathbf{u}_{\mathbf{m}}$ of prediction $\mathbf{m}$.
Based on the uncertainty map, we inject point loss $\mathcal{L}_p$ as a supplement to the common BCE loss $\mathcal{L}_b$, so the total loss of each decoder~(Inmodal Decoder, Amodal Decoder) is
\begin{equation}
    \mathcal{L}=\mathcal{L}_b+\mathcal{L}_p.
\end{equation}
Inspired by PointRend~\cite{kirillov2020pointrend}, we select only a certain number of points to exert point loss as shown in Fig.~\ref{fig:point}. First, $nK~(n>1)$ points are randomly chosen. By comparison, $cK~(0<c<1)$ points with relatively high uncertainty among these $nK$ points are adopted, and $(1-c)K$ points are picked stochastically. Consequently, a total of $K$ points will apply the point loss. The above $n,c,K$ are all constant parameters. For each selected point $\mathbf{x}$ in prediction $\mathbf{m}$, the point loss contains two terms:
\begin{equation}
    \mathcal{L}_p(\mathbf{x})=\alpha l_e(\mathbf{x})+\beta u_{\mathbf{m}}(\mathbf{x}).
\end{equation}
$l_e(\mathbf{x})$ is an extra BCE loss at full resolution aiming to increase the weights of these points in the original BCE loss. $\alpha$ is a constant coefficient to control the promotion level. $u_{\mathbf{m}}(\mathbf{x})$ is the uncertainty. High uncertainty means the predicted value is ambiguous, which makes it hard to select a threshold for ultimate binary classification. Therefore, a certain degree of punishment is imposed. $\beta$ is a constant coefficient to limit the penalty.
Through the point loss, challenging points with high uncertainty are given more attention and improved towards clearer predicted values, which is significant for the segmentation of the boundaries. On the other hand, the point sampling method enables the overall computation and memory to have only a slight increase.

Though elaborately designed, the previous modules are ineluctable to be puzzled at some details of the mask, thus we introduce a coarse-to-fine structure to enhance the quality. In previous modules, We have separately obtained the coarse inmodal prediction $\Tilde{\mathbf{m}}_v^i$ and the coarse amodal prediction $\Tilde{\mathbf{m}}_a^i$, which can fitly provide clues to each other. Therefore, we employ a refine module to fuse them and produce a more accurate amodal mask $\Tilde{\mathbf{M}}_a^i$. To explicitly indicate confusing points, we also inject the uncertainty map generated before into the module. That is, the refine module takes as input the concatenation of the original image $\mathbf{I}$, the coarse inmodal predictions $\Tilde{\mathbf{m}}_v^i$, the coarse amodal predictions $\Tilde{\mathbf{m}}_a^i$, and two corresponding uncertainty maps $\mathbf{u}_{\Tilde{\mathbf{m}}_v^i}, \mathbf{u}_{\Tilde{\mathbf{m}}_a^i}$. In practice, we select a simple yet effective convolutional network for the refine module contains only a ResNet block~\cite{he2016deep}:
\begin{equation}
    \Tilde{\mathbf{M}}_a^i=\mathrm{Conv}(\mathbf{I},\Tilde{\mathbf{m}}_v^i,\Tilde{\mathbf{m}}_a^i,\mathbf{u}_{\Tilde{\mathbf{m}}_v^i}, \mathbf{u}_{\Tilde{\mathbf{m}}_a^i}).
\end{equation}
When facing more complex datasets, this network can flexibly adjust to multiple blocks. 

\section{Experiments}
\label{sec:exp}
\subsection{Datasets and Metrics}
\paragraph{Datasets.}
In order to evaluate our proposed method, we conduct experiments on prominent datasets including KINS~\cite{qi2019amodal} and COCOA~\cite{follmann2019learning}. KINS and COCOA are commonly used datasets in the field of amodal segmentation, which are both based on real-world images with real occlusion.

The KINS dataset is built on the KITTI dataset~\cite{geiger2012we} with supplemental amodal annotations. KINS consists of 2 super-classes~(vehicle, person) and 7 sub-classes~(car, truck, pedestrian, \emph{etc}.) with 14991 images in total~(7474 images for training, 7517 images for testing). Covering objects with various occlusion ratios, KINS is the largest amodal street scene dataset.

The COCOA dataset is an extension of the Amodal COCO dataset~\cite{zhu2017semantic}. COCOA complements amodal annotations for a subset of the renowned COCO dataset~\cite{lin2014microsoft}. COCOA contains 80 categories~(dog, cat, apple, \emph{etc}.) and 3699 images~(2476 images for training, 1223 images for testing) in total.

\paragraph{Metrics.}
For we formulate the problem as a segmentation-only task, we choose mean intersection-over-union (IoU) as the metric, which is the ratio of intersecting pixels of the predicted mask and ground truth mask to their union. As in related work~\cite{gao2023coarse}, we adopt not only mean-IoU of the complete mask~($\text{mIoU}_{full}$), but also of the occluded region~($\text{mIoU}_{occ}$). $\text{mIoU}_{full}$ reflects the overall performance, while $\text{mIoU}_{occ}$ indicates the segmentation capability in our concerned occluded areas.

\subsection{Implementation Details}
PLUG is implemented based on the Pytorch framework~\cite{paszke2019pytorch}. Considering a balance of cost and effectiveness, we adopt the ViT-H version of pretrained SAM~\cite{kirillov2023segment,dosovitskiy2021an} with 636M parameters as the foundation and choose rank-16 LoRA adapters. Like related work~\cite{gao2023coarse}, we use double magnified visible bounding boxes to crop images. All input images are resized to $512\times 512$ and then processed. For constant parameters in the point loss, we select $\epsilon=3, n=4, c=0.75, K=256, \alpha=0.1, \beta=0.1$ in practice.
In the training, we utilize AdamW optimizer~\cite{loshchilov2018decoupled} and a dynamic learning rate, which reaches 0.001 after a quick 250-iteration warmup and then gradually decreases to 0. The $\beta_1,\beta_2$ and the weight decay of AdamW optimizer are set to 0.9, 0.999, 0.1. A whole training contains 50 epochs and is completed on 8 NVIDIA A100 Tensor Core GPUs taking about 5 hours for the COCOA dataset and 20 hours for the KINS dataset. In the testing, we set the threshold of logits to 0.3. Pixels exceeding the threshold are considered foreground, while other pixels are considered background.

\subsection{Comparison with Existing Methods}

\begin{table}[b]
\caption{The comparison of amodal segmentation performance on the KINS and COCOA datasets. The results of existing method are reported in C2F-Seg.}
\centering
\resizebox{0.8\textwidth}{!}{%
\begin{tabular}{c|c|cc|cc}
\midrule
\multirow{2}{*}{Methods} & \multirow{2}{*}{Venue}  & \multicolumn{2}{c|}{KINS}                  & \multicolumn{2}{c}{COCOA}                  \\ \cmidrule{3-6} 
          &             & $\ \text{mIoU}_{full}$ & $\text{mIoU}_{occ}$ & $\ \text{mIoU}_{full}$ & $\text{mIoU}_{occ}$ \\ \midrule
PCNet~\cite{zhan2020self} & CVPR'20         & 78.02                & 38.14               & 76.91                & 20.34               \\
VRSP~\cite{xiao2021amodal} & AAAI'21         & 80.70                & 47.33               & 78.98                & 22.92               \\
$\ $AISFormer~\cite{tran2022aisformer}$\ $ & $\ $BMCV'22$\ $  & 81.53                & 48.54               & 72.69                & 13.75               \\
C2F-Seg~\cite{gao2023coarse} & ICCV'23      & 82.22                & 53.60               & 80.28                & 27.71               \\ 
\midrule
SAM~(ViT-H)           & ICCV'23                & 63.42       & 18.57      & 74.15       & 14.36      \\
Ours~(ViT-B)           & -                & 86.91       & 59.74      & 81.43       & 29.67      \\
Ours~(ViT-L)           & -                & 88.10       & 61.42      & 83.23       & 32.88      \\
\rowcolor{gray!10}
Ours~(ViT-H)           & -                & \textbf{88.85}       & \textbf{62.66}      & \textbf{83.71}       & \textbf{34.12}      \\ \midrule
\end{tabular}
}
\label{tab:compare}
\end{table}

\paragraph{Baselines.}
For better evaluation of our proposed method, we choose several typical and recent approaches for comparison, which are PCNet~\cite{zhan2020self}, VRSP~\cite{xiao2021amodal}, AISFormer~\cite{tran2022aisformer}, and C2F-Seg~\cite{gao2023coarse}. These baselines are all designed for the amodal segmentation task, among which C2F-Seg is the state-of-the-art method now. It should be noted that the amodal segmentation task is different from the amodal completion task. The latter takes the ground truth visible masks of all objects as input, which provide a lot more information. In the amodal segmentation task, visible masks are also obtained by prediction. Therefore, our comparative experiments are conducted within the scope of amodal segmentation. For fair comparison, we use pre-detected visible masks in accordance with related work~\cite{gao2023coarse}~(by AISFormer on KINS, by VRSP on COCOA) to calculate corresponding bounding boxes, which will be taken as input of the model.

\paragraph{Results.}
As shown in Table~\ref{tab:compare}, our PLUG approach significantly outperforms existing methods with large margins on both KINS and COCOA datasets, reaching state-of-the-art performance. On the KINS dataset, PLUG achieves prime performance 6.63/9.06 higher than C2F-Seg. 
On the more challenging COCOA dataset, PLUG still performs well and attains 83.71/34.12 on mean IoU, which beats C2F-Seg by 2.51/4.31.
For better performance, the largest ViT-H version (645M parameters in total) of PLUG is recommended.
Without changing other settings, replacing foundation SAM with a smaller version will result in some performance degradation. Even so, the ViT-L version~(316M parameters in total) and the ViT-B version~(96M parameters in total) of PLUG, which are smaller in size than C2F-Seg~(382M parameters in total), can still surpass the performance of C2F-Seg.
Furthermore, we can view the foundation models as infrastructure due to their numerous functionalities. 
In this perspective, our PLUG model possesses a very lightweight size. The incremental trainable model contains only 12.7M parameters.
When multiple incremental models are built together on the foundation model, it will actually save costs compared to specific models.

\begin{figure}[t]
  \centering
   \includegraphics[width=0.88\linewidth]{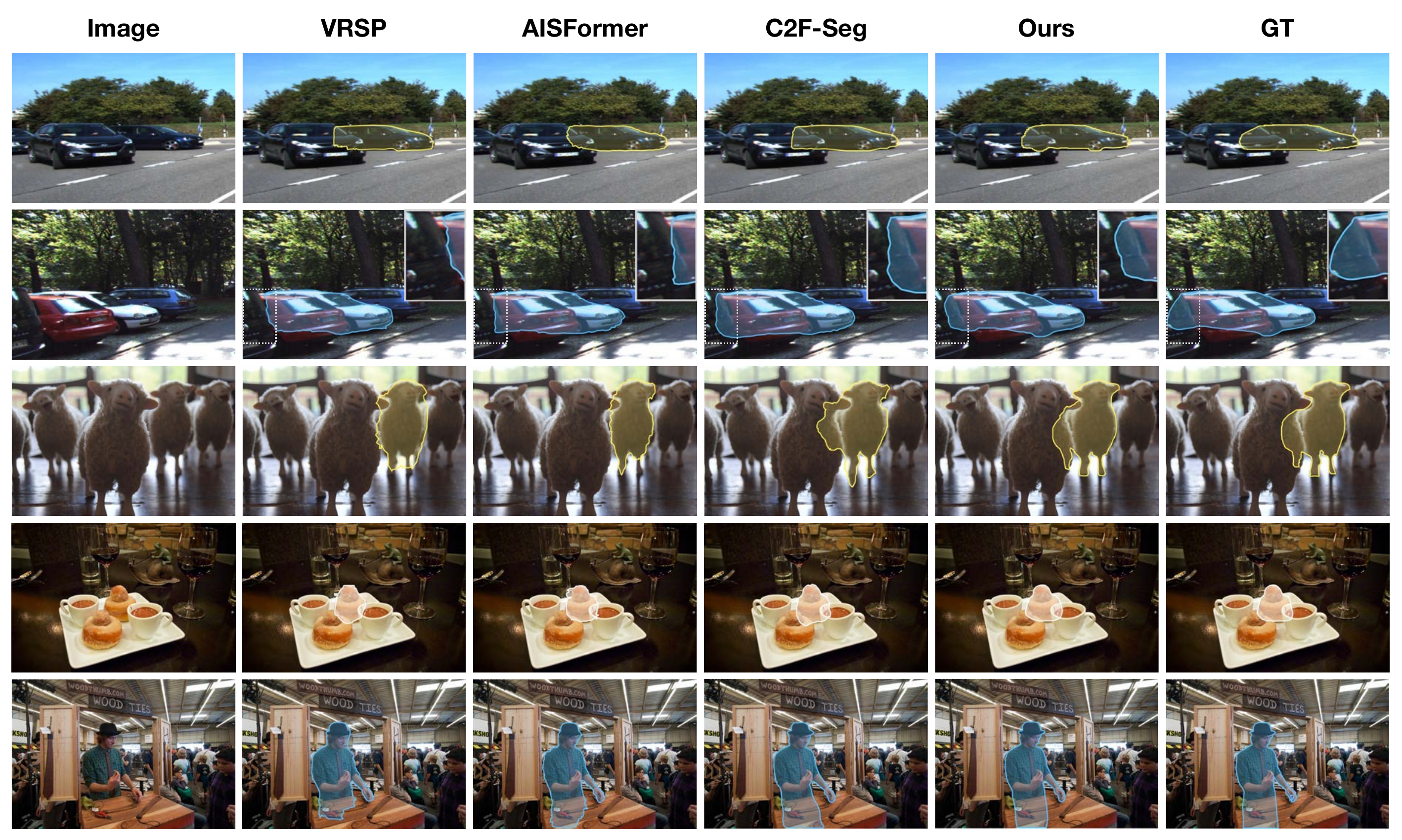}
   \caption{\textbf{Qualitative results.} The qualitative comparison of predicted amodal masks from VRSP, AISFormer, C2F-Seg and our proposed PLUG approach. The first two rows are from the KINS dataset, while the last three rows are from the COCOA dataset. Zoom in for a better view.}
   \label{fig:comparison}
\end{figure}

Several qualitative results are also shown in Fig.~\ref{fig:comparison}, which intuitively display our superiority in performance. In contrast with other approaches, PLUG can provide more accurate and complete shapes. For the car in the first row, the tires and the curved rear can be identified in our prediction, while other methods only provide a rough range. For the sheep in the third row, our prediction includes its hind leg, which is not covered by other approaches.

\subsection{Ablation Study}
As shown in Table~\ref{tab:ablation}, we conduct ablation study on both KINS and COCOA datasets, in which we compare pretrained SAM with models that sequentially add fine-tuning, point loss, refine module, and parallel LoRA on it.
Although we input amodal bounding boxes, the pretrained SAM~(at the 1st row) can hardly provide valid predictions in the occluded regions.
The 2nd row utilizes plain LoRA to finetune SAM. Consequently, the performance acquires a dramatic improvement and almost equals the state-of-the-art C2F-Seg approach. On this basis, our proposed method brings more advance.

\begin{table}[h]
\caption{The ablation study on the KINS and COCOA datasets. FT, PT, RM, PL represent fine-tuning, point loss, refine module, and parallel LoRA. $\mathbf{I}, \mathbf{B}_v,\mathbf{B}_a$ represent the original image, visible and amodal bounding box, respectively. All experiments are based on the ViT-H version of SAM.}
\centering
\resizebox{0.85\textwidth}{!}{%
\begin{tabular}{c|ccccc|cc|cc}
\midrule
\multirow{2}{*}{} & \multicolumn{5}{c|}{Settings}  & \multicolumn{2}{c|}{KINS}                  & \multicolumn{2}{c}{COCOA}                  \\ \cmidrule{2-10} 
                  & $\ $FT & PT & RM & PL & Input                       & $\ \text{mIoU}_{full}$ & $\text{mIoU}_{occ}$ & $\ \text{mIoU}_{full}$ & $\text{mIoU}_{occ}$ \\ \midrule
$\ $1$\ $                 & & & & & $\mathbf{I,B}_a\ $          & 63.42                & 18.57               & 74.15                & 14.36               \\
$\ $2$\ $                 &\checkmark & & & & $\mathbf{I,B}_v$              & 82.47                & 50.10               & 80.16                & 27.07               \\
$\ $3$\ $                 &\checkmark &\checkmark & & & $\mathbf{I,B}_v$              & 84.73                & 53.45               & 81.88                & 29.35               \\
$\ $4$\ $                 &\checkmark &\checkmark &\checkmark & & $\mathbf{I,B}_v$           & 85.48                & 55.12               & 82.56                & 31.28               \\
\rowcolor{gray!10}
$\ $5$\ $                 &\checkmark &\checkmark &\checkmark &\checkmark & $\mathbf{I,B}_v$           & \textbf{88.85}                & \textbf{62.66}               & \textbf{83.71}                & \textbf{34.12}               \\
\midrule
\end{tabular}
}
\label{tab:ablation}
\end{table}

\paragraph{The Effect of Point Loss.}
To verify the benefit of the point loss, we can compare the 2nd and 3rd rows in Table~\ref{tab:ablation}. By adding the point loss, the performance shows a significant promotion of 2.26/3.35 on KINS and 1.72/2.28 on COCOA. 
To confirm the point loss does not incur excessive computation, we also record the training speed on the KINS dataset. It takes about 4.3~seconds per iteration for the final version, while it takes about 3.9~seconds per iteration if the point loss is removed. The results reflect that the impact of point loss on training duration is limited~(10\% or so).

\paragraph{The Effect of Refine Module.}
To demonstrate the importance of the refine module, we can compare the 3rd and 4th rows in Table~\ref{tab:ablation}. The outcome shows that the latter notably beats the former by 0.75/1.67 on the KINS dataset and 0.68/1.93 on the COCOA dataset. 

\paragraph{The Effect of Parallel LoRA.}
In Table~\ref{tab:ablation}, we employ only a plain single-branch LoRA at the 2nd to 4th rows, while applying the two-branch parallel LoRA structure at the 5th row. To evaluate the effectiveness of parallel LoRA, we can compare the 4th and 5th rows that vary only in the LoRA structure. The latter wins a remarkable advantage of 3.37/7.54 on KINS and 1.15/2.84 on COCOA.

\begin{table}[t]
\caption{The comparison of different ranks of LoRA. ``Params'' means the parameter quantities of incremental trainable models. All experiments are based on the ViT-H version of SAM.}
\centering
\resizebox{0.68\textwidth}{!}{%
\begin{tabular}{c|c|cc|cc}
\midrule
\multirow{2}{*}{$\ $Rank$\ $} & \multirow{2}{*}{$\ $Params$\ $} & \multicolumn{2}{c|}{KINS}                  & \multicolumn{2}{c}{COCOA}                  \\ \cmidrule{3-6} 
             &            & $\ \text{mIoU}_{full}$ & $\text{mIoU}_{occ}$ & $\ \text{mIoU}_{full}$ & $\text{mIoU}_{occ}$ \\ \midrule
4  & 8.8~M         &      88.01           &    61.23            &      83.44           &   33.25             \\
8  & 10.0~M         &      88.32           &    61.87            &      83.60           &   33.81             \\
\rowcolor{gray!10}
16 & 12.7~M      & \textbf{88.85}                & \textbf{62.66}               & \textbf{83.71}                & \textbf{34.12}               \\
32 & 17.9~M          &      88.54           &    62.08            &      83.54           &   33.78             \\
\midrule
\end{tabular}
}
\label{tab:rank}
\end{table}

\paragraph{The Effect of The Rank of LoRA Adapters.}
Due to the rank $r$ of LoRA adapters possesses a direct correlation with parameter quantity and final performance~\cite{hu2022lora}, we make a careful choice based on experimental results. As shown in Table~\ref{tab:rank}, our selected $r=16$ has a compact size and achieves better performance compared to other attempts. 
\begin{figure}[b]
  \centering
   \includegraphics[width=0.54\linewidth]{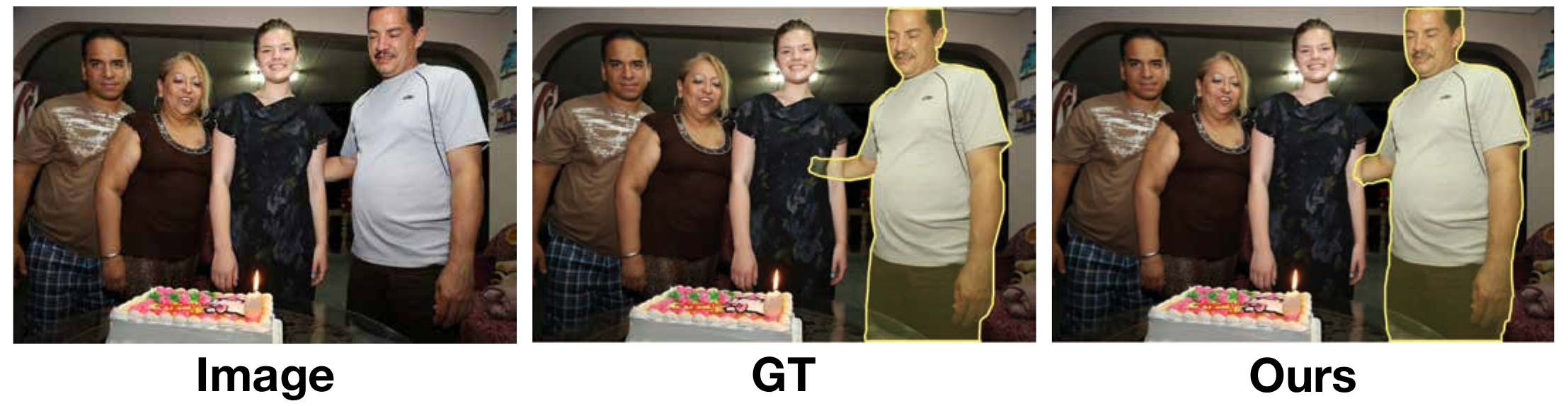}
   \caption{\textbf{An example of the limitation.} In this image, the man's occluded arm is not well segmented.}
   \label{fig:limitation}
\end{figure}
\section{Limitation}
As shown in Fig.~\ref{fig:limitation}, we find that our model is \emph{not} competent to give valid predictions for some non-rigid instances like people in special poses, which actually reflects a common problem in existing amodal segmentation methods that the diversity interpretations of hidden parts are not effectively addressed. Is it practicable to introduce prompts of more efficacious modalities or directly provide multiple reasonable predictions? We leave this challenge to future work.

\section{Conclusion}
\label{sec:conclusion}
In this work, we revisit the amodal segmentation task and propose a brand-new SAM-based framework, PLUG. Introducing the parallel LoRA structure to avoid mutual disturbance of different regions and the concept of uncertainty to identify ambiguous points, PLUG establishes a novel hierarchical focus mechanism to well utilize the superb capabilities of the foundation model SAM and serve the amodal segmentation task. Experimental results demonstrate PLUG significantly outperforms all existing methods and achieves state-of-the-art performance.

%
%
\bibliographystyle{plain}
\bibliography{main}
\end{document}